%% file: icm-arxiv.tex
\documentclass{article}

\usepackage[utf8]{inputenc}
\pdfoutput=1
\usepackage[sc,osf]{mathpazo}
\usepackage[margin=1.1in,letterpaper]{geometry}
\usepackage[sort&compress,numbers]{natbib}
\usepackage[colorlinks=true,citecolor=blue,breaklinks]{hyperref}
\usepackage[hyphenbreaks]{breakurl} 

\usepackage{caption}

\makeatletter
\newlength\aftertitskip     \newlength\beforetitskip
\newlength\interauthorskip  \newlength\aftermaketitskip

\def\maketitle{\par
 \begingroup
   \def\thefootnote{\fnsymbol{footnote}}
   \def\@makefnmark{\hbox to 4pt{$^{\@thefnmark}$\hss}}
   \@maketitle \@thanks
 \endgroup
\setcounter{footnote}{0}
 \let\maketitle\relax \let\@maketitle\relax
 \gdef\@thanks{}\gdef\@author{}\gdef\@title{}\let\thanks\relax}

\def\@startauthor{\noindent \normalsize\bf}
\def\@endauthor{}
\def\@starteditor{\noindent \small {\bf Editor:~}}
\def\@endeditor{\normalsize}
\def\@maketitle{\vbox{\hsize\textwidth
 \linewidth\hsize \vskip \beforetitskip
 {\begin{center} \LARGE\@title \par \end{center}} \vskip \aftertitskip
 {\def\and{\unskip\enspace{\rm and}\enspace}%
  \def\addr{\small\it}%
  \def\email{\hfill\small\tt}%
  \def\name{\normalsize\bf}%
  \def\AND{\@endauthor\rm\hss \vskip \interauthorskip \@startauthor}
  \@startauthor \@author \@endauthor}
}}

\makeatother

\pdfoutput=1                    

\usepackage{url}            
\usepackage{booktabs}       
\usepackage{amsfonts}       
\usepackage{nicefrac}       
\usepackage{microtype}      

\usepackage{graphicx}
\usepackage{subfigure}
\usepackage{booktabs} 

\usepackage{amsmath}
\usepackage{framed}
\usepackage{amssymb}
\usepackage{array}
\usepackage{amsthm}
\usepackage{verbatim} 
\usepackage{enumerate}
\usepackage{wrapfig}
\usepackage{amsbsy}
\usepackage{float}
\usepackage{algorithm}
\usepackage{tabularx}
\usepackage{listings}
\usepackage{xcolor}
\usepackage{enumitem}

\usepackage{natbib}
\usepackage{xcolor}

\newcommand{\G}{\mathcal{G}}
\newcommand{\D}{\mathcal{D}}
\newcommand{\Ds}{\mathcal{D}}
\newcommand{\R}{\mathbb{R}}

\newcommand{\Ps}{\mathcal{P}}
\newcommand{\Qs}{\mathcal{Q}}
\newcommand{\Fs}{\mathcal{F}}
\newcommand{\Fh}{\widehat{F}}
\newcommand{\Xs}{\mathcal{X}}
\newcommand{\Ys}{\mathcal{Y}}
\newcommand{\Ns}{\mathcal{N}}
\newcommand{\Cs}{\mathcal{C}}

\newcommand{\Js}{\mathcal{J}}
\newcommand{\As}{\mathcal{A}}

\newcommand{\risk}{\mathcal{R}}
\newcommand{\erisk}{\widehat{\mathcal{R}}}

\newcommand{\rad}{\widehat{\mathfrak{R}}}  
\newcommand{\yb}{\mathbf{y}}

\newcommand{\fagg}{f_{\text{Agg}}}
\newcommand{\fup}{f_{\text{Up}}}
\newcommand{\fread}{f_{\text{Read}}}

\newcommand{\mo}{\{\!\!\{} 
\newcommand{\mc}{\}\!\!\}}

\newcommand{\hash}{\text{\textsc{Hash}}}

\newcommand{\nla}{\phi_1}
\newcommand{\nlb}{\phi_2}
\newcommand{\tree}{\mathcal{T}}

\newcommand{\Gs}{\mathcal{G}}
\newcommand{\bell}{\mathrm{b}} 

\DeclareMathOperator{\tr}{tr}
\DeclareMathOperator{\supp}{supp}

\newtheorem{theorem}{Theorem}

\newtheorem{proposition}{Proposition}
\newtheorem{corollary}{Corollary}
\newtheorem{definition}{Definition}

\numberwithin{equation}{section}

\begin{document}

\title{Theory of Graph Neural Networks: Representation and Learning}

\author{  \name Stefanie Jegelka \email{stefje@mit.edu}\\
  \addr{Dept.\ of EECS and CSAIL, MIT} 
}

\maketitle

\begin{abstract}
  Graph Neural Networks (GNNs), neural network architectures targeted to learning representations of graphs, have become a popular learning model for prediction tasks on nodes, graphs and configurations of points, with wide success in practice. This article summarizes a selection of the emerging theoretical results on approximation and learning properties of widely used message passing GNNs and higher-order GNNs, focusing on representation, generalization and extrapolation. Along the way, it summarizes mathematical connections. \footnote{This article is a modified version of an article originally written for the International Congress of Mathematicians 2022.} 
\end{abstract}


\input{intro2}

\input{representation2}

\input{generalization1}

\input{extrapolation1}

\input{disc}


\subsection*{Acknowledgments}
The author would like to thank Keyulu Xu, Derek Lim, Behrooz Tahmasebi, Vikas Garg, Tommi Jaakkola, Andreas Loukas, Jingling Li, Mozhi Zhang, Simon Du, Ken-ichi Kawarabayashi, Weihua Hu, Jure Leskovec, Joan Bruna and Yusu Wang for discussions on the theory of GNNs, collaborations and pointers.

This work was partially supported by NSF CAREER award 1553284, NSF SCALE MoDL award 2134108, and NSF CCF-2112665 (TILOS AI Research Institute).


\bibliographystyle{plainnat}
\bibliography{allrefs}

\end{document}

%% file: intro2.tex
\section{Introduction}\label{sec:intro}
There has been growing interest in solving machine learning tasks when the input data is given in form of a graph $G = (V,E,X,W)$ from a set of attributed graphs $\Gs$, where $X \in \R^{d \times |V|}$ contains vectorial attributes for each node, and $W \in \R^{d_w \times |E|}$ contains attributes for each edge ($X$ and $W$ may be empty). Examples include predictions in social networks, recommender systems and link predicton (given two nodes, predict an edge), property prediction of molecules, prediction of drug interactions, traffic prediction, forecasting physics simulations, and learning combinatorial optimization algorithms for hard problems \citep{duvenaud15b,grover16,battaglia2016interaction,lesong17,hamilton17s,zitnik18,sato19,sanchez20,cappart2021survey,derrow21} (and many other references). These examples use two types of tasks: (1) given a graph $G$, predict a label $F(G)$; (2) given a graph $G$ and node $v \in V(G)$ (or pair of nodes ($u,v$)), predict a node label $f(v)$.

Solving these tasks demands a sufficiently rich \emph{embedding} of the graph or of each node that captures structural properties as well as the attribute information. While graph embeddings have been a widely studied topic, including spectral embeddings and graph kernels, recently, \emph{Graph Neural Networks (GNNs)}\cite{gori05,scarselli09,merkwirth05,gilmer17,kipf17,hamilton17} have emerged as an empirically broadly successful model class that, as opposed to, e.g., spectral embeddings, allows to adapt the embedding to the task at hand, generalizes to other graphs of the same input type, and incorporates attributes. Due to space limits, this survey focuses on the popular message passing (spatial) GNNs, formally defined below, and a selection of their rich mathematical connections, with an excursion into higher-order GNNs.

When \emph{learning} a GNN, we observe $N$ i.i.d.\ samples $\D = \{G_i, y_i\}_{i=1}^N \in (\Gs \times \Ys)^N$ drawn from an underlying distribution $\Ps$ on $\Gs \times \Ys$. The \emph{labels} $y_i$ are often given by an unknwon \emph{target function} $g(G_i)$, and observed with or without i.i.d.\ noise. Given a (convex) loss function $\ell: \Gs \times \Ys \times \Ys \to \R$ that measures prediction error, i.e., mismatch of $y$ and $F(G)$, such as the squared loss or cross-entropy, we aim to estimate a model $F$ from our GNN model class $\Fs$ to minimize the expected loss (\emph{population risk}) $\risk(F)$:
\begin{equation}
  \label{eq:risk}
  \min_{F \in \Fs}\; \mathbb{E}_{(G,y) \sim \Ps}[ \ell( G,y, F(G)) ] \equiv \min_{F \in \Fs}\; \risk(F).
\end{equation}
When analyzing learning and risk, three main questions become important:

\textbf{1. Representational power (Section~\ref{sec:representation}).} Which target functions $g$ can be approximated well by a GNN model class $\Fs$? Answers to this question relate to graph isomorphism testing, approximation theory for neural networks, local algorithms and representing invariance/equivariance under permutations. 

\textbf{2. Generalization (Section~\ref{sec:generalization}).} Even with sufficient approximation power, we can only \emph{estimate} a function $\Fh \in \Fs$ from $\Ds$: since $\Ps$ and hence $\risk(F)$ is not accessible, the common learning or \emph{training} procedure is to minimize the \emph{empirical risk} $\erisk(F)$:
\begin{equation}
  \label{eq:emprisk}
  \Fh \in \arg\min_{F \in \Fs}\; \frac1N \sum_{i=1}^N \ell( G_i ,y_i, F(G_i)) \equiv \arg\min_{F \in \Fs}\; \erisk(F).
\end{equation}
\emph{Generalization} asks how well $\Fh$ is performing according to the population risk, i.e., $\risk(\Fh)$, as a function of $N$ and model properties. Good generalization may demand explicit (e.g., via penalties) or implicit regularization (e.g., via the optimization algorithm, typically variants of stochastic gradient descent). Hence, generalization analyses involve the complexity of the model class $\Fs$, the target function, the data and the optimization procedure.

\textbf{3. Generalization under distribution shifts (Section~\ref{sec:extrapolation}).} In practice, a learned model $\Fh$ is often deployed on data from a distribution $\Qs \neq \Ps$, e.g., graphs of different size, degree or attribute ranges, for instance, $\supp(Q) \supset \supp(P)$. In which cases can we expect successful \emph{extrapolation} to $\Qs$? This depends on the structure of the graphs and the task, formalizable via graph limits, local structures and algorithmic structures, e.g., dynamic programming.

Beyond these topics, GNNs have close connections to graph signal processing as learnable filters, geometric learning, and probabilistic inference. The first and third connection are covered in \citep{hamilton20book}.

Lastly, a disclaimer: GNNs are a rapidly evolving research area. Hence, it is almost impossible to include all possible works, and any survey necessarily misses some. The author apologizes in advance for any references that were not covered.

\subsection{Graph Neural Networks (GNNs)}\label{sec:gnn}
In this article, we focus on \emph{Message passing graph neural networks (MPNNs)}, which follow an iterative scheme \cite{gori05,scarselli09,merkwirth05,gilmer17,kipf17,hamilton17}. Throughout, they maintain a representation (embedding) $h^{(t)}_v \in \R^{d_t}$ for each node $v \in V$. 
In each iteration $t$, they update each embedding $h^{(t)}_v$ as a function of its neighbors' embeddings and possible edge attributes:
\begin{align}
  h_v^{(0)} &= x_v, \quad\quad \forall v \in V& \text{(Initialization)}\\
  m_v^{(t)} &= \fagg^{(t)}\big(h^{(t-1)}_v,\, \mo h^{(t-1)}_u, w(u,v) \mid u \in \Ns(v) \mc\big), \; 1 \leq t < T \qquad &\text{(Aggregate)}\\
  \label{eq:upd}
  h_v^{(t)} &= \fup(h_v^{(t)} , m_v^{(t)}) \qquad\quad &\text{(Update)}.
\end{align}
The final node representation $f(v) = h_v^{(T)}, \forall v \in V$ is the last iterate, possibly concatenated with a linear classifier. Throughout this paper, $\Ns(v) \subset V$ denotes the neighborhood of $v \in V$, and $\mo \cdot \mc$ a multiset. Here, $h^{(t)}_v$ encodes the $t$-hop neighborhood of node $v$, i.e., the subgraph of all nodes reachable from $v$ within $t$ steps. The number of iterations $T$ is also termed the GNN \emph{depth}, and one iteration may be viewed as a layer.

The \emph{aggregation function} $\fagg^{(t)}: \R^{d_{t-1}} \to \R^{d_{t}}$ plays a major role and is shared by all nodes within an iteration. It is a nonlinear function of the form
\begin{align}\label{eq:agg}
  \fagg^{(t)}\Big(h^{(t-1)}_v, \mo h^{(t-1)}_u, w(u,v) \mid u \in \Ns(v) \mc\Big) = \nla^{(t)} \Big( \sum_{u \in \Ns(v)} \nlb^{(t)}\big(h_u^{(t-1)},h_v^{(t-1)}, w(u,v)\big)\Big).
\end{align}
The sum may also be replaced by an average, degree-normalized sum or coordinate-wise min or max.
In the most general form, the functions $\nla, \nlb$ are 
implemented as \emph{multi-layer perceptrons (MLPs)}, neural networks that alternate linear transformations and coordinate-wise nonlinear activations such as the ReLU ($\sigma(a) = \max\{a,0\}$) or sigmoid function ($\sigma(a) = (1+\exp(-a))^{-1}$):
\begin{align}
  \mathrm{MLP}(h ; \theta) &=  \sigma(W^{(M)} \ldots     \sigma(W^{(2)} \sigma(W^{(1)}h + b^{(1)}) + b^{(2)}) \ldots + b^{(M)}).
\end{align}
The learnable parameters $\theta$ of the MLP are the weight matrices $W^{(j)}$ and bias vectors $b^{(j)}$.
The \emph{update} $\fup$ in Equation~\eqref{eq:upd} is typically a weighted combination with learnable weight matrices:
\begin{align}\label{eq:update}
  \fup\Big( h_v^{(t)} , m_v^{(t)}\Big) &= \sigma\Big( W^{(t)}_1 h_v^{(t)} + W^{(t)}_2 m_v^{(t)}\Big) \quad \text{ or }\quad \fup\Big(h_v^{(t)} , m_v^{(t)}\Big) = m_v^{(t)}.
\end{align}
Finally, if a graph-level prediction is desired, all node representations can be aggregated by a permutation invariant \emph{readout} function
\begin{align}
  F(G) = \fread\big( \mo h_v^{(T)} \mid v \in V \mc\big).
\end{align}
Here, we assume the readout has the form~\eqref{eq:agg} or is a simple sum or average. Typically, all parameters are learned jointly via stochastic gradient descent minimizing the empirical risk.

Throughout this article, $n=|V|$ denotes the number of nodes and $N$ the number of training data points.

\textbf{Permutation invariance.} An important property of GNNs is permutation invariance of the graph, and equivariance of the node representations. 
Let $A \in \R^{n \times n}$ be the adjacency matrix of a graph $G \in \G$, and $X \in \R^{n \times d}$ its node features. Permutation invariance/equivariance means that for all permutation matrices $P \in \R^{n \times n}$ and all $G \in \Gs$:
\begin{align}
  F(PAP^\top, PX) &= F(A, X)\\
  f(PAP^\top, PX, v) &= f(A, X, v). 
\end{align}

\textbf{Spectral GNNs.} Besides message passing GNNs, other architectures have been devised. One important example are \emph{spectral} graph neural networks \cite{bruna14,defferrard16}, which learn a function of the graph Laplacian $L$, i.e., $F(G) = p(L)X = V p(\Lambda) V^\top X$, where $V$ and $\Lambda$ are the matrices of eigenvectors and eigenvalues (diagonal matrix), respectively, and $p$ is a polynomial. In the sequel, we will focus on message passing GNNs.

%% file: representation2.tex
\section{Representational power of GNNs}\label{sec:representation}
For functions on graphs, representational power has mainly been studied in terms of graph isomorphism: which graphs a GNN can distinguish. Via variations of the Stone-Weierstrass theorem, these results yield universal approximation results. Other works bound the ability of GNNs to compute specific polynomials of the adjacency matrix and to distinguish graphons \cite{dehmamy19,magner20}.
 Observed limitations of MPNNs have inspired higher-order GNNs (Section~\ref{subsec:higher}). Moreover, if all node attributes are unique, then analogies to local algorithms yield algorithmic approximation results and lower bounds (Section~\ref{subsec:nodeids}).

\subsection{GNNs and Graph isomorphism testing}\label{sec:mpgnn_iso}
A standard characterization of the discriminative power of GNNs is via the hierarchy of the \emph{Weisfeiler-Leman (WL)} algorithm for graph isomorphism testing, also known as color refinement or vertex classification \cite{read77}, which was inspired by the work of Weisfeiler and Leman \cite{weisfeiler68,weisfeiler76}. The WL algorithm does not entirely solve the graph ismomorphism problem, but its power has been widely studied.

A \emph{labeled} graph is a graph endowed with a node coloring $l: V(G) \to \Sigma$ for some sufficiently large alphabet $\Sigma$. Given a labeled graph $(G,l)$, the 1-dimensional WL algorithm (1-WL) iteratively computes a node coloring $c_l^{(t)}: V(G) \to \Sigma$. Starting with $c_l^{(0)}$ in iteration $t=0$, in iteration $t>0$ it sets for all $v \in V$
\begin{equation}\label{eq:WLcol}
  c_l^{(t)}(v) = \hash\left( c^{t-1}_{l}(v),\; \mo c^{t-1}_{l}(u) \mid u \in \Ns(v)\mc \right),
\end{equation}
where \textsc{Hash} is an injective map from the input pair to $\Sigma$, i.e., it assigns a unique color to each neighborhood pattern. To compare two graphs $G, G'$, the algorithm compares the multisets $\mo c_l^{(t)}(v) \mid v \in V(G) \mc$ and $\mo c_l^{(t)}(u) \mid u \in V(G') \mc$
in each iteration. If the sets differ, then it determines that $G \neq G'$. Otherwise, it terminates when the number of colors in iteration $t$ and $t-1$ are the same, which occurs after at most $\max\{|V(G)|,|V(G')|\}$ iterations.

The computational analogy between the 1-WL algorithm and MPNNs is obvious. Since the WL algorithm uniquely colors each neighborhood, the coloring $c_l^{(t)}(v)$ always \emph{refines} the coloring $h^{(t)}_v$ from a GNN.
\begin{theorem}[\cite{xuhlj19,morris19}]\label{thm:WLupper}\label{thm:WLinjective}
  If for two graphs $G,G'$ a message passing GNN outputs $f_G(G) \neq f_G(G')$, then the 1-WL algorithm will determine that $G \neq G'$.\\
  For any $t$, there exists an MPGNN such that $c_l^{(t)} \equiv h^{(t)}$. A sufficient condition is that the aggregate, update and readout operations are injective multiset functions.
\end{theorem}
GNNs that use the degree for normalization in the aggregation \cite{kipf17} can be equivalent to the 1-WL agorithm too, but with one more iteration in the WL algorithm \cite{geerts21}.

\subsubsection{Representing multiset functions}
Theorem~\ref{thm:WLinjective} demands the neighbor aggregation $\fagg$ to be an injective multiset function on sets $S$ ($|S| \leq M$). Theorem~\ref{thm:univ_multiset} shows how to universally approximate multiset functions.
\begin{theorem}[\cite{xuhlj19,wagstaff19}]\label{thm:univ_multiset}
  Any multiset function $G: \Xs^{\leq M} \to \R$ on a countable domain $\Xs$ can be expressed as
  \begin{equation}
    G(S) = \nla\Big( \sum\nolimits_{s \in S} \nlb(s)\Big),
  \end{equation}
  where $\nla: \R^{d_1} \to \R^{d_2}$ and $\nlb: \R^{d_2} \to \R$ are nonlinear functions.
\end{theorem}
The proof idea is to show that there exists an injective function of the form $\sum_{s \in S} \phi(s)$. 
The above result is an extension of a universal approximation result for set functions \cite{zaheer17,qi17,ravanbakhsh16}, and suggests a neural network model for sets where $\nla, \nlb$ are approximated by MLPs. 
The Graph Isomorphism Network (GIN) \cite{xuhlj19} implements this sum decomposition in the aggregation function to ensure the ability of injective operations.

Here, the latent dimension $d_2$ plays a role. Proofs for countable domains use a discontinuous mapping $\nla$ into a fixed-dimensional space, whereas MLPs universally approximate \emph{continuous} functions \cite{cybenko89}.
Continuous set functions on $\R^{\leq M}$ (i.e., set cardinality $|S| \leq M$) can be sum-decomposed as above with continuous $\nla, \nlb$ and latent dimension at least $d_2 = M$. This dimension is a necessary and sufficient condition for universal approximation \cite{wagstaff19}. For GNNs, this means $d_2$ must be at least the maximum degree $\deg(G)$ of the input graph $G$.


\subsubsection{Implications for graph distinction}
Theorem~\ref{thm:WLupper} allows to directly transfer any known result for the 1-WL test to MPNNs. For instance, the 1-WL test succeeds to distinguish graphs sampled uniformly from all graphs on $n$ nodes with high probability, and failure probability going to zero as $n \to \infty$ \cite{babai79,babai80}. 1-WL can also distinguish any non-isomorphic pair of trees \cite{immerman90}. It fails for regular graphs, as all node colors will be the same. The graphs that the 1-WL algorithm can distinguish from any non-isomorphic graph can be recognized in quasi-linear time \cite{arvind15}.
See also \cite{cai89,kiefer15,arvind15} for more detailed results on the expressive power of variants of the WL algorithm.

\subsubsection{Computation trees and structural graph properties}\label{sec:comptrees}
\begin{figure}
  \includegraphics[width=0.9\textwidth]{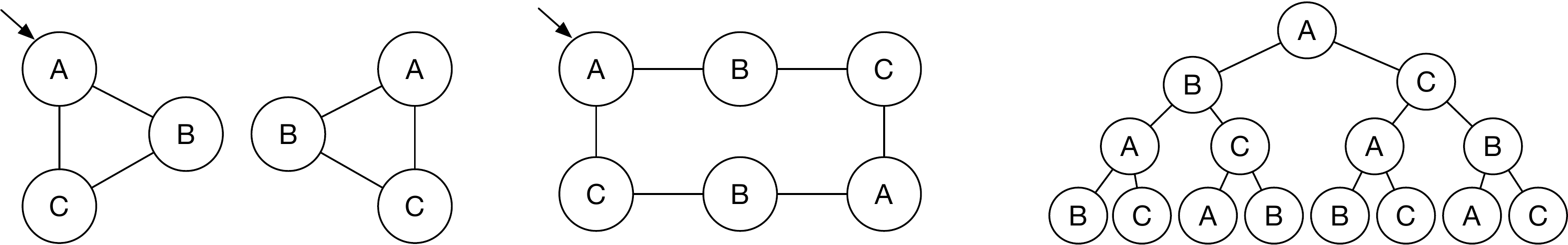}
  \caption{Graphs $G_1$ (left, 2 connected components) and $G_2$ (middle) with node attributes indicated by letters. The computation tree rooted at the node with arrow (right) agrees in both graphs, and likewise for the other nodes. Hence, 1-WL and MPNNs cannot distinguish $G_1$ and $G_2$. Figure adapted from \cite{garg20}. \label{fig:comptree}}
\end{figure}

To further illustrate the implications of GNNs' discriminative power, we look at some specific examples. The maximum information contained in any embedding $h_v^{(t)}$ can be characterized by a computation tree $\tree(h^{(t)}_v)$, i.e., an ``unrolling'' of the message passing procedure. The 1-WL test essentially colors computation trees. 
The tree  $\tree(h^{(t)}_v)$ is constructed recursively: let $\tree(h^{(0)}_v) = x_v$ for all $v \in V$. For $t > 0$, construct a root with label $x_v$ and, for any $u \in \Ns(v)$ construct a child subtree $\tree(h^{(t-1)}_u)$. Figure~\ref{fig:comptree} illustrates an example.
\begin{proposition}
If for two nodes $u \neq v$, we have $\tree(h^{(t)}_v) = \tree(h^{(t)}_u)$, then $h^{(t)}_v = h^{(t)}_u$.
\end{proposition}
Comparing computation trees directly implies that MPNNs cannot distinguish regular graphs. It also shows further limitations with practical impact, as indicated in Figure~\ref{fig:comptree}, in particular for learning combinatorial algorithms and for predicting properties of molecules, where functional groups are of key importance. We say a class of models $\Fs$ \emph{decides} a graph property if there exists an $F \in \Fs$ such that for any two $G, G'$ that differ in the property, we obtain $F(G) \neq F(G')$.
\begin{proposition}\label{prop:struct-prop}
  MPNNs cannot decide girth, circumference, diameter, radius, existence of a conjoint cycle, total number of cycles, and existence of a $k$-clique \cite{garg20}.
  MPNNs cannot count induced (attributed) subgraphs for any connected pattern of 3 or more nodes, except star-shaped patterns \cite{chen20count}.
\end{proposition}
Motivated by these limitations, generalizations of GNNs were proposed that provably increase their representational power. Two main directions are to (1) introduce node IDs (Section~\ref{subsec:nodeids}), and (2) use higher-order functions that act on tuples of nodes (Section~\ref{subsec:higher}).
%

\subsection{Node IDs, local algorithms, combinatorial optimization and lower bounds}\label{subsec:nodeids}
The major weaknesses of MPNNs arise from their inability to identify nodes as the origin of specific messages. Hence, MPNNs can be strengthened by making nodes more distinguishable. The gained representational power follows from connections with local algorithms, where the input graph defines both the computational problem and the network topology of a distributed system: there, each node $v \in V$ is a local machine and generates a local output, and all nodes execute the same algorithm, without faults.

\textbf{Approximation Algorithms.} Sato et al.~\cite{sato19} achieve a partial node distinction by transferring the idea of a \emph{port numbering} from local algorithms. Edges incident to each node are numbered as outgoing ports. In each round, each node simultaneously sends a message to each port, but the messages can differ across ports: 
\begin{equation}
  m_v^{(t)} = f^{(t)}_{\text{Agg}}\left( \mo (\text{port}(u,v), \text{port}(v,u), h_u^{(t-1)} )  \mc \right).
\end{equation}
Permutation invariance, though, is not immediate. This corresponds to the vector-vector consistent (VV$_C$) model for local algorithms \cite{hella12}.
The VV$_C$ analogy allows to transfer results on representing approximation algorithms. CPNGNN is a specific VV$_C$ GNN model. 
\begin{theorem}[\cite{sato19}]
  There exists a CPNGNN that can compute a $(\deg(G)+1)$-approximation for the minimum dominating set problem, a CPNGNN that can compute a 2-approximation for the minimum vertex cover problem, but no CPNGNN can do better. No CPNGNN can compute a constant-factor approximation for the maximum matching problem.
\end{theorem}
Adding a weak vertex 2-coloring leads to further results. Despite the increased power compared to MPNN, CPNGNNs retain most limitations of Proposition~\ref{prop:struct-prop} \cite{garg20}.

A more powerful alternative is to endow nodes with fully unique identifiers \cite{sato21,loukas20}. E.g., augmenting the GIN model (the most expressive MPNN) \cite{xuhlj19} with random node identifiers yields a model that can decide subgraphs that MPNN and CPNGNN cannot \cite{sato21}. This model can further achieve better approximation results for minimum dominating set ($H(\deg(G)+1) + \epsilon$), where $H$ is the harmonic number) and maximum matching ($1+\epsilon$).


\textbf{Turing completeness.} Analogies to local algorithms further imply that MPNNs with unique node IDs are \emph{Turing complete}, i.e., they can compute any function that a Turing machine can compute, including graph isomorphism. In particular, the proof shows an equivalence to the Turing universal LOCAL model from distributed computing \cite{anglouin80,linial92,naor93}.
\begin{theorem}[\cite{loukas20}]\label{thm:turing}
  If $\fup$ and $\fagg$ are Turing complete functions and the message passing GNN gets unique node IDs, then the classes GNN and LOCAL are equivalent. For any MPNN $F$ there exists a local algorithm $\mathcal{A}$ of the same depth, such that $F(G) = \mathcal{A}(G)$, and vice versa.
\end{theorem}
\begin{corollary}[\cite{loukas20}]
  Under the conditions in Theorem~\ref{thm:turing}, if the GNN depth (number of iterations) is at least $\mathrm{diameter}(G)$ and the width (dimensionality of the embeddings $h^{(t)}_v$) is unbounded, then MPNNs can compute any Turing computable function over connected attributed graphs.
\end{corollary}
\textbf{Lower bounds.} 
For bounded size, GNNs lose computational power. Via analogies to the CONGEST model \cite{peleg00}, which bounds message sizes, one can transfer results on decision, optimization and estimation problems on graphs. These lead to lower bounds on the product of depth and width of the GNN if the nodes do not have access to a random generator. Here, as before, \emph{width} of a GNN refers to the dimensionality of the embeddings $h^{(t)}_v$.
\begin{theorem}[\cite{loukas20}]\label{thm:congest}
  If a problem cannot be solved in less than $T$ rounds in CONGEST using messages of at most $b$ bits, then it cannot be solved by an MPNN of width $w \leq (b-\log_2 n)/p = O(b/\log n)$ and depth $T$, where $p= \Theta(n)$.
\end{theorem}
Theorem~\ref{thm:congest} directly implies lower bounds for solving combinatorial problems, e.g., $Tw = \Omega(n/\log n)$ for cycle detection and computing diameter, and $T\sqrt{w} = \Omega(\sqrt{n}/\log n)$ for minimum spanning tree, minimum cut and shortest path \cite{loukas20}.


Moreover, we can transfer ideas from communication complexity. The \emph{communication capacity} $c_f$ of an MPNN $f$ (with unique node IDs) is the maximumm number of symbols that the MPNN can transmit between any two disjoint sets $V_1, V_2 \subset V$ of nodes when viewed as a communication network:
$c_f \leq \mathrm{cut}(V_1,V_2) \sum_{t=1}^T \min\{m_t, w_t\} + \sum_{t=1}^T \gamma_t$, where $T$ is the GNN depth, $w_t$ the width of layer $t$, $m_{t}$ the size of the messages, and $\gamma_t$ the size of a global state that is maintained. The communication capacity of the MPNN must be at least $c_f = \Omega(n)$ to distinguish all trees, and $c_f = \Omega(n^2)$ to distinguish all graphs \cite{loukas20dist}. 
By relating discrimination and function approximation (Section~\ref{subsec:approximation}), these results have implications for function approximation, too.

\textbf{Random node IDs.} While unique node IDs are powerful in theory, in many practical examples the input graphs do not have unique IDs. An alternative is to assign random node IDs \cite{dasoulas20,abboud21}. This can still yield GNNs that are essentially permutation invariant: while their outputs are random, the outputs for different graphs are still sufficiently separated \cite{abboud21}. This leads to a probabilistic universal approximation result: 
\begin{theorem}[\cite{abboud21}]
  Let $g: \Gs \to \R$ be a permutation invariant function on graphs of size $n \geq 1$. Then for all $\epsilon, \delta > 0$ there exists an MPNN $F$ with access to a global readout and with random node IDs such that for every $G \in \Gs$ it holds that $\Pr( |F(G) - g(G)| \leq \epsilon) \geq 1-\delta$.
\end{theorem}
The proof builds on a result by \cite{barcelo20} that states that any logical sentence in FOC$_2$ can be expressed by the addressed GNN. The logic considered here is a fragment of first-order (FO) predicate logic that allows to incorporate counting quantifiers of the form $\exists^{\geq k} x \psi(x)$, i.e., there are at least $k$ elements $x$ satisfying $\psi$, but is restricted to two variables. FOC$_2$ is tightly linked with the 1-WL test: for any nodes $u,v \in V$ in any graph, 1-WL colors $u$ and $v$ the same if and only if they are classified the same by all FOC$_2$ classifiers \cite{cai92}. 

Other approaches include \cite{you21identity,vignac20structural}.


\paragraph{Augmentations.} Another successful idea is to augment node attribute vectors with attributes that contain structural or topological information \cite{zhao20,graphSNNanon22,bouritsas20,liu20count,monti18motif,ying21transformer,dwivedi21gentransf,zhang2020graphbert,lim22}.

\subsection{Higher-order GNNs}\label{subsec:higher}

Instead of adding unique node IDs, one may increase the expressive power of GNNs by encoding subsets of $V$ that are larger than the single nodes used in MPNNs. Three such directions are: (1) neural network versions of higher-dimensional WL algorithms, (2) (non)linear equivariant operations, and (3) recursion. Other strategies that could not be covered here use, e.g., simplicial and cell complexes~\cite{bodnar21,bodnar21topo}. 

Most of these GNNs act on $k$-tuples $s \in V^k$ and 
may be written in a unified form via tensors $H^{(t)} \in R^{n^k \times d_t}$, where the first $k$ coordinates index the tuple, and $H^{(t)}_{s,:} \in \R^{d_t}$ is the representation of tuple $s$ in layer $t$.
For MPNNs, which use node and edge information, $H^{(0)} \in \R^{n \times n \times (d+1)}$. The first $d$ channels of $H^{(0)}$ encode the node attributes: $H^{(0)}_{v,v,1:d} = x_v$ and $H^{(0)}_{u,v,1:d} = 0$ for $u \neq v$. The final channel captures the adjacency matrix $A$ of the graph: $H^{(0)}_{:,:,(d+1)} = A$.
Node embeddings are computed by a permutation equivariant network $f: \R^{n^k \times (d_0)} \to \R^{n^k \times (d_T)}$:
\begin{equation}\label{eq:tensor-gnn}
  f(G) = m \circ S_E \circ F^{(T)} \circ \ldots \circ F^{(1)} \circ \text{\textsc{shape}}(G),
\end{equation}
where $m: \R^{d_T} \to \R^{d_{\text{out}}}$ is an MLP that is applied to each representation $h^T_s$ separately, $S_E: \R^{n^k \times d_T} \to \R^{n \times d_T}$ is a reduction $S_E(H)_{v,:} = \sum_{s \in V^k: s_1 = v} H_{s,:}$, and each layer $F^{(t)}: \R^{n^k \times d_{t-1}} \to \R^{n^k \times d_{t}}$ is a message passing (aggregation and update) operation for MPNNs, and will be defined for higher-order networks.
The first operation shapes the input into the correct tensor form, if needed. For a graph embedding, we switch to a reduction $S_I: \R^{n^k \times d_T} \to \R^{d_T}$, $S_I(H) = \sum_{s \in V^k} H_{s,:}$ and apply the MLP $m$ to the resulting vector: $F(G) = m \circ S_I \circ F^{(T)} \circ \ldots \circ F^{(1)} \circ \text{\textsc{shape}(G)}.$ Different GNN models differ in their layers $F^{(t)}$, which must be permutation equivariant.

\subsubsection{Higher-order WL networks}
Extending analogies between MPNNs and the 1-WL algorithm \cite{xuhlj19,morris19}, the first class of higher-order GNNs imitates versions of the $k$-dimensional WL algorithm. The $k$-WL algorithms are defined on $k$-tuples of nodes, and different versions differ in their aggregation and definition of neighborhood. In iteration $0$, the \emph{$k$-WL} algorithm labels each $k$-tuple $s \in V^k$ by a unique ID for its isomorphism type. Then it aggregates over neighborhoods $\Ns^{\text{WL}}_i(s) = \{ (s_1, s_2, \ldots s_{i-1}, v, s_{i+1}, \ldots s_k) \mid \forall v \in V\}$ for $1 \leq i \leq k$:
\begin{align}
  c_i^{(t)}(s) &= \mo c^{(t-1)}(s') \mid s' \in \Ns^{\text{WL}}_i(s)\mc\;\;\quad 1 \leq i \leq k, s \in V^k\\
  c^{(t)}(s) &= \hash\big( c^{(t-1)}(s),\, c_1^{(t)}(s), c_2^{(t)}(s) \ldots, c_k^{(t)}(s)\big).
\end{align}
For two graphs $G,G'$ the $k$-WL algorithm then decides ``not isomorphic'' if $\mo c^{(t)}(s) \mid s \in V(G)^k\mc \neq \mo c^{(t)}(s') \mid s' \in V(G')^k\mc$ for some $t$, and returns ``maybe isomorphic'' otherwise. Like the 1-WL test, the $k$-WL test decides ``not isomorphic'' only if $G \ncong G'$.
The \emph{Folklore $k$-WL} algorithm ($k$-FWL) differs in its update rule, which ``swaps'' the order of the aggregation steps \cite{cai92}: 
\begin{align}
  c_u^{(t)}(s) &= \big( c_{(u, s_2, \ldots, s_k)}^{(t-1)}, c_{(s_1, u, s_3, \ldots, s_k)}^{(t-1)}, \ldots, c_{(s_1, \ldots, s_{k-1},u)}^{(t-1)}\big),\; \forall u \in V, s \in V^k \\
  c^{(t)}(s) &= \hash\big( c^{(t-1)}(s), \mo  c_u^{(t)}(s) \mid u \in V\mc  \big), \;\; \forall s \in V^k.
\end{align}
The 1-WL and 2-WL tests are equivalent, and for $k \geq 2$, the $(k+1)$-WL test can distinguish strictly more graphs than the $k$-WL test \cite{cai92}. The $k$-FWL algorithm is as powerful as the $(k+1)$-WL algorithm for $k \geq 2$ \cite{grohe15}.

\textbf{Set-WL GNN.} Since computations on $k$-tuples are expensive, \citet{morris19} consider a GNN that corresponds to a set version of a $k$-WL algorithm. For any \emph{set} $S \subseteq V$ with $|S|=k$, let $\Ns^{\text{set}}(S) = \{ T \subset V, |T| = k \mid |S \cap T| = k-1\}$. The set-based WL ($k$-SWL) algorithm then updates as
\begin{align}
  c^{(t)}(S) = \hash\Big( c^{(t-1)}(S), \mo c^{(t-1)}(T) \mid T \in \Ns^{\text{set}}(S)\mc \Big);
\end{align}
its GNN analogue uses the aggregation and update (cf.~Eqns.~\eqref{eq:agg},\eqref{eq:update})
\begin{equation}\label{eq:kgnn}
  h_S^{(t+1)} = \sigma\Big( W_1^{(t)}h_S^{(t)} + \sum\nolimits_{T \in \Ns^{\text{set}}(S)} W^{(t)}_2h_T^{(t)} \Big),
\end{equation}
where $\sigma$ is a coordinatewise nonlinearity (e.g., sigmoid or ReLU). 
This family of GNNs is equivalent in power to the $k$-SWL test \cite{morris19} (Theorem~\ref{thm:disc-joint}).
For computational efficiency, a local version restricts the neighborhood of $S$ to sets $T$ such that the nodes $\{u,v\} = S \Delta T$ in the symmetric difference are connected in the graph. This local version is weaker \cite{abboud21}.


\textbf{Folklore WL GNN.} In analogy to the $k$-FWL algorithm, Maron et al.~\cite{maron19powerful} define $k$-FGNNs with aggregations
\begin{equation}\label{eq:kfgnn}
  h_s^{(t)} = f^{(t)}_{\text{Up}}\Big( h^{(t-1)}_s, \sum_{v \in V} \prod_{i=1}^k f^{(t)}_i(h^{(t-1)}_{(s_1, \ldots, s_{i-1}, v, s_{i+1}, \ldots s_k)} ) \Big).
\end{equation}
For $k=2$, this model can be implemented via matrix multiplications. The input to the aggregation, for all pairs of nodes simultaneously, is a tensor $H \in \R^{n \times n \times d_t}$, with $H_{(u,v),:} = h_{(u,v)}$. The initial $H^{(0)} \in \R^{n \times n \times (d+1)}$ is defined as in the beginning of Section~\ref{subsec:higher}.

To compute the aggregation layer, first, we apply three MLPs $m_1, m_2: \R^{d_1} \to \R^{d_2}$ and $m_3: \R^{d_1} \to \R^{d_3}$ to each embedding $h_{(u,v)}$ in $H$: $m_l(H)_{(u,v),:} = m_l(H_{(u,v),:})$ for $1 \leq l \leq 3$. Then one computes an intermediate representation $H' \in \R^{n \times n \times d_2}$ by multiplying matching ``slices'' of the outputs of $m_1, m_2$: $H'_{:,:,i} = m_1(H)_{:,:,i} \cdot m_2(H)_{:,:,i}$. The final output of the aggregation is the concatenation $(m_3(H),H') \in \R^{n \times n \times (d_2 + d_3)}$. 
%
%
A variation of this model, a low-rank global \emph{attention} model, was shown to relate attention and the 2-FWL algorithm via \emph{algorithmic alignment}, which we discuss in Section~\ref{sec:gen_align} \cite{puny20}. Attention in neural networks introduces learned pair-wise weights in the aggregation function.

The family of $k$-FGNNs is a class of nonlinear equivariant networks, and is equivalent in power to the $k$-FWL test and the $(k+1)$-WL test \cite{maron19powerful,azizian21} (Theorem~\ref{thm:disc-joint}).


\subsubsection{Linear equivariant layers.}
While the models discussed so far rely on message passing, the GNN definition~\eqref{eq:tensor-gnn} only requires permutation equivariant or invariant operations in each layer. The $k$-linear (equivariant) GNNs ($k$-LEGNNs), introduced in \cite{maron19}, allow more general linear equivariant operations.
In $k$-LEGNNs, each layer $F^{(t)} = \sigma \circ L^{(t)}: \R^{n^k \times d_{t-1}} \to \R^{n^k \times d_{t}}$ is a concatenation of a linear equivariant function $L^{(t)}$ and a coordinate-wise nonlinear activation function. The function $\sigma$ may also be replaced with a nonlinear function $f_1^{(t)}: \R^{d_{t+1/2}} \to \R^{d_{t+1}}$ (an MLP) applied separately to each tuple embedding $L^{(t)}(H^{(t-1)})_{s,:}$.

Characterizations of equivariant functions or networks were studied in \cite{kondor18,hartford18,ravanbakhsh17,kondor18compact}. Maron et al.~\cite{maron19} explicitly characterize all invariant and equivariant linear layers, and show that the vector space of linear invariant or equivariant functions $f: \R^{n^k} \to \R^{n^\ell}$ 
has dimension 
 $\bell(k)$ and $\bell(k+\ell)$, respectively, where $\bell(k)$ is the $k$-th Bell number. When including multiple channels and bias terms, one obtains the following bounds.
\begin{theorem}[\cite{maron19}]
 The space of invariant (equivariant) linear layers $\R^{n^k \times d} \to \R^{d'}$ ($\R^{n^k \times d} \to \R^{n^k \times d'}$ )
has dimension $dd' \bell(k) + d'$ (for equivariant: $dd' \bell(2k) + d'\bell(k)$).
\end{theorem}
The associated GNN model uses one parameter (coefficient) for each basis tensor. Importantly, the number of parameters is independent of the number of nodes.
The proof for identifying the basis tensors sets up a fixed point equation with Kronecker products of any permutation matrix that any equivariant tensor must satisfy. The solutions to these equations are defined by equivalence classes of multi-indices in $[n]^k$. Each equivalence class is represented by a partition $\gamma$ of $[k]$, e.g., $\gamma = \{ \{1\}, \{2,3\}\}$ includes all multi-indices $(i_1,i_2,i_3)$ where $i_1 \neq i_2,i_3$ and $i_2 = i_3$. The basis tensors $B^\gamma \in \{0,1\}^{n^{k}}$ are then such that $B^\gamma_s = 1$ if and only if $s \in \gamma$.

Linear equivariant GNNs of order $k$ ($k$-LEGNNs) parameterized with the full basis are as discriminative as the $k$-WL algorithm \cite{maron19powerful} (Theorem~\ref{thm:disc-joint}).
To achieve this discriminative power, each entry $H^{(0)}_{s,:}$ in the input tensor encodes an initial coloring of the isomorphism type of the subgraph indexed by the $k$-tuple $s$.

\subsubsection{Summary of Representational Power via WL} The following theorem summarizes equivalence results between the GNNs discussed so far and variants of the WL test. Following \cite{azizian21}, we here use equivalence relations, as they suffice for universal approximation in Section~\ref{subsec:approximation}. For a set $\Fs$ of functions defined on $\Gs$, define an equivalence relation $\rho$ via the joint discriminative power of all functions $F \in \Fs$, i.e., for any two graphs $G,G' \in \Gs$:
\begin{equation}
  (G,G') \in \rho(\Fs)\; \Leftrightarrow\; \forall F \in \Fs,\; F(G) = F(G').
\end{equation}
\begin{theorem}\label{thm:disc-joint}
  The above GNN families have the following equivalences:
  \begin{align}
  \rho(\text{MGNN}) &= \rho(\text{2-WL}) 
  \quad \text{\cite{xuhlj19}}\\
  \rho(k\text{-set-GNN}) &= \rho(\text{$k$-SWL}) \quad \text{\cite{morris19}}\\
  \label{eq:disc-equiv}
  \rho(\text{$k$-LEGNN}) &= \rho(\text{$k$-WL}) 
  \quad \text{\cite{maron19,geerts20}} \\
  \rho(\text{$k$-FGNN}) &= \rho(\text{$(k$+$1)$-WL}) 
  \quad \text{\cite{maron19powerful,azizian21}}.
\end{align}
\end{theorem}
Analogous results hold for equivariant models (for node representations), with the exception of equality \eqref{eq:disc-equiv}, which becomes an inclusion: $\rho(\text{$k$-LGNN}_E) \subseteq \rho(\text{$k$-WL}_E)$ \cite{azizian21}.

\subsubsection{Relational Pooling.}
One option to obtain nonlinear permutation invariant functions is to average permutation-sensitive functions over the permutation group $\Pi_{n}$. Murphy et al.~\cite{murphy19,murphy19jan} propose such a model. 
Concretely, if $A \in \R^{n \times n}$ denotes the adjacency matrix of the input graph $G$ and $X \in \R^{n \times d}$ the matrix of node attributes, then
\begin{equation}\label{eq:relpool}
  F_{\text{RP}}(G) = \frac{1}{n!}\sum_{\pi \in \Pi_{n}} g(A_{\pi,\pi},X_{\pi}) = g(\pi \cdot H^{(0)}),
\end{equation}
where $X_{\pi}$ is $X$ with permuted rows, and $H^{(0)}$ is the tensor combining adjacency matrix and node attributes. Here, $g$ is any permutation-sensitive function, and may be modeled via various nonlinear function approximators, e.g. neural networks such as fully connected networks (MLPs), recurrent neural networks or a combination of a convolutional network applied to $A$ and an MLP applied to $X$. In particular, this model allows to implement graph isomorphism testing via node IDs (cf.\ Section~\ref{subsec:nodeids}) if $g$ is a universal approximator \cite{murphy19}. For instance, node IDs may be permuted over nodes and concatenated with the node attributes:
\begin{equation}
  F_{\text{RP}}(G) = \frac{1}{n!}\sum_{\pi \in \Pi_{n}}(A_{\pi,\pi},[X^\top_{\pi}, I_n]^\top) = \frac{1}{n!}\sum_{\pi \in \Pi_{n}}g(A,[X^\top, (I_n)_{\pi}]^\top),
\end{equation}
where $I_n\in \R^{n \times n}$ is the identity matrix. If $g$ is an MPNN, the resulting model is strictly more powerful than the 1-WL test and hence $g$ by itself.

The drawback of the \emph{Relational Pooling} \eqref{eq:relpool} is its computational intractability. Various approximations have been considered, e.g., defining canonical orders, stochastic approximations, and applying $g$ to all possible $k$-subsets of $V$. In the latter case, increasing $k$ strictly increases the expressive power.
\emph{Local Relational Pooling} is a variant that applies relational pooling to the $k$-hop subgraphs centered at each node, and then aggregates the results. This operation provably allows to identify and count subgraphs of size up to $k$ \cite{chen20count}.

\subsubsection{Recursion}\label{subsec:recursion}
A general strategy for encoding a graph is to encode a collection of subgraphs and then aggregate these encodings. The question of what graphs $G,G'$ this process allows to distinguish ($F(G) \neq F(G')$), depends on the collection of subgraphs used, the subgraph encoding function and the aggregation function. As a special case, this process includes the reconstruction hypothesis \cite{kelly1957,ulam60}, i.e., the question whether any graph $G$ can be reconstructed from the collection of its subgraphs $G\setminus \{v\}$, for all $v$ in $G$. 
One challenge with the reconstruction hypothesis is that no alignment of the $G\setminus \{v\}$ is available. Indeed, node correspondences across subgraphs provide important information \cite{tahmasebi21,bevilacqua21}.

Indeed, the expressive power of a model based on subgraph encodings depends on the set of subgraphs, the type of subgraph encodings and the aggregation. Tahmasebi et al.~\cite{tahmasebi21} use recursion as a powerful tool: instead of iterative message passing or layering, a \emph{recursive} application of the above subgraph embedding step, even with a simple set aggregation like~\eqref{eq:agg}, can enable a GNN that can count any bounded-size subgraphs, as opposed to MPNNs (Prop.~\ref{prop:struct-prop}).

Let $\mathcal{N}_r(v)$ be the $r$-hop neighborhood of $v$ in $G$. \emph{Recursive neighborhood pooling (RNP)} encodes \emph{intersections} of such neighborhoods of different radii. Given an input graph $G$ with node attributes $( h^{\text{in}}_u)_{u \in V(G)}$ and a sequence $(r_1, \ldots, r_{t})$ of radii, RNP-GNN recursively encodes the node-deleted $r_1$-neighborhoods $G_v = \mathcal{N}_{r_1}(v)\setminus \{v\}$ of all nodes $v \in V$ after marking the deletion in augmented representations $h^{\text{aug}}_{u}$, $u \in V$. It then combines the results, and returns node representations of all nodes. Concretely, for each $v \in V$, it computes $G_v$ and 
\begin{align}
h^{\text{aug}}_{u} &= (h^{\text{in}}_u, \mathbf{1}[ (u,v) \in E(G_v)]);\;\,\; \forall u \in G_v\\
  ({h}'_{v,u})_{u \in G_v} &\gets \textsc{RNP-GNN}\big(G_v, (h^{\text{aug}}_{u} )_{u \in G_v}, (r_2,r_3,\ldots,r_t)\big) \quad \text{(recursion)}\\
\text{return}\qquad  h^{\text{out}}_v &= \fagg^{(t)} \Big ( h_v^{\text{in}}, \mo h'_{v,u}: u \in  G_v\mc \Big),\;\; \forall v \in V \quad\quad \text{(aggregation)}.
\end{align}
If the sequence of radii is empty (base case), then the algorithm returns the input attributes $h^{\text{in}}_u$. In contrast to \emph{iterative} message passing, the encoded subgraphs here correspond to intersections of local neighborhoods. Together with the node deletions and markings that retain node correspondences, this maintains more structural information. Formally, if the sequence of radii dominates a covering sequence for a subgraph $H$ of interest, then, with appropriate parameters, RNP can count the induced and non-induced subgraphs of $G$ isomorphic to $H$ \cite{tahmasebi21}. The computational cost is $O(n^k)$ for recursion depth $k$, and better for very sparse graphs, in line with computational lower bounds.

\subsection{Universal approximation}\label{subsec:approximation}
Distinguishing given graphs is closely tied to approximating continuous 
functions on graphs. In early work, Scarselli et al.~\cite{scarselli09approx}  take a fixed point view and show a universal approximation result for infinite-depth MPNNs whose layers are contraction operators, for functions on equivalence classes defined by computation trees. Dehmamy et al.~\cite{dehmamy19} analyze the ability of GNNs to compute polynomials of the adjacency matrix.

Later works derive universal approximation results for graph and permutation-equivariant functions from graph discrimination results via extensions of the Stone-Weierstrass theorem \cite{azizian21,keriven19,maron19univ,chen19}.
\citet{maron19univ} argue that $H$-invariant networks (for a permutation group $H$) can universally approximate $H$-invariant polynomials, which in turn can universally approximate any invariant function \cite{yarotsky21}. Keriven and Peyr\'e \cite{keriven19} do not fix the size of the graph and show that shallow equivariant networks can, with a single set of parameters, well approximate a function on graphs of varying size. Both constructions involve very large tensors.

More generally, a Stone-Weierstrass theorem (for symmetries) allows to translate Theorem~\ref{thm:disc-joint} into universal approximation results. Let $\Cs_I(\Xs,\Ys)$ 
be the set of invariant continuous functions from $\Xs$ to $\Ys$. Then a class $\Fs$ of GNNs is \emph{universal} if its closure $\overline{\Fs}$ (in uniform norm) on a compact set $K$ is the entire $\Cs_I(K,\R^p)$.
\begin{theorem}[\cite{azizian21}]
Let $K_{\text{disc}} \subseteq \Gs_n \times \R^{d_0\times n}$, $K \subseteq \R^{d_0\times n}$ be compact sets, where $\Gs_n$ is the set of all unweighted graphs on $n$ nodes.
\begin{align}
  \overline{\text{MGNN}} &= \{ f \in \Cs_I(K_{disc},\R^p) : \rho(\text{$2$-WL}) \subseteq \rho(f)\}\\
  \overline{\text{$k$-LEGNN}} &= \{ f \in \Cs_I(K,\R^p) : \rho(\text{$k$-WL}) \subseteq \rho(f)\}\\
  \overline{\text{$k$-FGNN}} &= \{ f \in \Cs_I(K,\R^p) : \rho(\text{$(k$+$1)$-WL}) \subseteq \rho(f)\}.
\end{align}
Analogous relations hold for equivariant functions, except for $\overline{\text{$k$-LEGNN}_E} = \{ f \in \Cs_E(K,\R^{n\times p}) : \rho(\text{$k$-LEGNN}_E) \subseteq \rho(f)\}$, which is a superset of $\{ f \in \Cs_E(K,\R^{n\times p}) : \rho(\text{$k$-WL}_E \subseteq \rho(f)\}$.
\end{theorem}

%% file: generalization1.tex
\section{Generalization}\label{sec:generalization}

Beyond approximation power, a second important question in machine learning is generalization. 
\emph{Generalization} asks how well the estimated function $\Fh$ is performing according to the population risk, i.e., $\risk(\Fh)$, as a function of the number of data points $N$ and model properties. Good generalization may demand explicit (e.g., via a penalty term) or implicit regularization (e.g., via the optimization algorithm). Hence, generalization analyses involve aspects of the complexity of the model class $\Fs$, the target function we aim to learn, the data and the optimization procedure. This is particularly challenging for neural networks, due to the nested functional form and the non-convexity of the empirical risk.

A classic learning theoretic perspective bounds the \emph{generalization gap} $\risk(\Fh) - \erisk(\Fh)$ via the complexity of the model class $\Fs$ (Section~\ref{sec:gen_comp}). These approaches do not take into account possible implicit regularization via the optimization procedure. One possibility to do so is via the \emph{Neural Tangent Kernel} approximation (Section~\ref{sec:gen_gntk}). Finally, for more complex, structured target functions, e.g., algorithms or physics simulations, one may want to also consider the structure of the target task. One such option is  \emph{Algorithmic Alignment} (Section~\ref{sec:gen_align}). Another strategy for obtaining generalization bounds is via \emph{algorithmic stability}, the condition that, if one data point is replaced, the outcome of the learning algorithm does not change much. This strategy led to some early bounds for spectral GNNs \cite{verma19}.

\subsection{Generalization bounds via complexity of the model class}\label{sec:gen_comp}

\textbf{Vapnik-Chervonenkis dimension.} The first GNN generalization bound was based on bounding the Vapnik-Chervonenkis (VC) dimension \cite{vapnik71} 
of the GNN function class $\Fs$. The \emph{VC dimension} of $\Fs$ expresses the maximum cardinality of a set of data points such that for any binary labeling of the data, some GNN in $\Fs$ can perfectly fit, i.e., \emph{shatter}, the set. The VC dimension directly leads to a bound on the generalization gap. Here, we only state the results for sigmoid activation functions.
\begin{theorem}[\cite{scarselli18vc}]\label{thm:vcdim}
  The VC dimension of GNNs with $p$ parameters, $H$ hidden neurons (in the MLP) and input graphs of size $n$ is $O(p^2H^2n^2)$.
\end{theorem}
Strictly speaking, Theorem~\ref{thm:vcdim} is for node classification with one hidden layer in the aggregation function MLPs.
The VC dimension directly yields a bound on the generalization gap: for a class $\Fs$ with VC dimension $D$, with probability $1-\delta$, it holds that
\begin{equation}
  \risk(\Fh) - \erisk(\Fh) \leq O\left(\sqrt{\frac{D}{N}\log \frac{N}{D}}\right) + \sqrt{\frac{1}{2N}\log \frac{1}{\delta}}.
\end{equation}
Interestingly, in these bounds, GNNs are a generalization of recurrent neural networks 
\cite{scarselli18vc}. 
The VC dimension bounds for GNNs are the same as for recurrent neural networks \cite{koiran97}; the bounds for fully connected MLPs are missing the factor $n^2$ \cite{karpinski97}.

\textbf{Rademacher Complexity.} Bounds that are in many cases tighter can be obtained via Rademacher complexity. The \emph{empirical Rademacher complexity} $\rad_S(\Fs)$ of a function class $\Fs$ measures how well it can fit ``noise'' in the form of uniform random variables $\sigma = (\sigma_1, \ldots, \sigma_N)$ in $\{-1, +1\}$:
\begin{equation}
  \rad_S(\Fs) = \mathbb{E}_{\sigma}\left[\sup_{F \in \Fs} \frac{1}{N} \sum_{i=1}^N \sigma_i F(x_i)\right],
\end{equation}
  for a fixed data sample $S = \{x_1, \ldots, x_N\}$. Similarly to VC dimension, $\rad_S(\Fs)$ provides a bound on the probability of error under the full data distribution: $\mathbb{P}[\text{error}(F)] \leq \erisk(F) + 2\rad_S(\Js) + 3\sqrt{\frac{\log(2/\delta)}{2N}}$, where $\Js$ is the class of functions $F \in \Fs$ concatenated with the loss.
  Garg et al.~\cite{garg20} analyze a GNN that applies a logistic linear binary classifier at each node, averages these predictions for a graph-level prediction, and uses a \emph{mean field update} \cite{dai15}: $h^{(t)}_v = \phi(W_1 x_v + W_2 \rho( \sum_{u \in N(v)} g(h^{(t-1)}_u)))$, where $\phi, \rho, g$ are nonlinear functions with bounded Lipschitz constant that are zero at zero (e.g., tanh), and $\|W_1\|_F, \|W_2\|_F \leq B$. The logistic classifier outputs a ``probability'' for the label 1, and is evaluated by a margin loss function that gives a (scaled) penalty if the ``probability'' of the correct label is below a threshold ($\frac{\gamma+1}{2}$).
  \begin{theorem}[\cite{garg20}]\label{thm:rademacher}
    Let $\mathcal{C}$ be the product of the Lipschitz constants of $\phi, \rho, g$ and $B$; $T$ the number of GNN iterations; $w$ the dimension of the embeddings $h^{(t)}_v$, and $d$ the maximum branching factor in the computation tree. Then the generalization gap of the GNN can be bounded as: $\widetilde{O}(\frac{wd}{\sqrt{N}\gamma})$ for $\mathcal{C} < 1/d$, $\widetilde{O}(\frac{wdT}{\sqrt{N}\gamma})$ for $\mathcal{C} = 1/d$ and $\widetilde{O}(\frac{wd\sqrt{wT}}{\sqrt{N}\gamma})$ for $\mathcal{C} > 1/d$.
  \end{theorem}
  The factor $d$ is equal to $\max_{v \in G}\deg(v)-1$. For recurrent neural networks, the same bounds hold, but with $d=1$ \cite{chen20rnn}: a sequence is a tree with branching factor 1.
  To compare these bounds with the bounds based on VC-dimension, we use that $H=w$, $n > d$, and $p$ the size of the matrices $W$, i.e., about $w^2$, and obtain a VC-dimension based generalization bound of $\tilde{O}(w^3n/\sqrt{N})$, ignoring log factors.
Later work tightens the bounds in Theorem~\ref{thm:rademacher} by using a \emph{PAC-Bayesian} approach \cite{liao21}.

\subsection{Generalization bounds via the Neural Tangent Kernel}\label{sec:gen_gntk}
Infinitely wide neural networks can be related to kernel learning techniques via the \emph{Neural Tangent Kernel} \cite{jacot18,du19global,du19over,arora19exact,arora19finegrained}. Du et al.~\cite{du19gntk} extend this analysis to a broad class of GNNs. The main idea underlying the Neural Tangent Kernel (NTK) is to approximate a neural network $F(\theta, G)$ with a kernel derived from the training dynamics.
Assume we fit $F(\theta, G)$ with the squared loss $L(\theta) = \sum_{i=1}^N\ell(F(\theta, G_i), y_i) = \frac12 \sum_{i=1}^N(F(\theta, G_i) - y_i)^2$, where $\theta \in \R^m$ collects all parameters of the network. If we optimize with gradient descent with infinitesimally small step size, i.e., $\frac{d\theta(t)}{dt} = -\nabla L(\theta(t))$, then the network outputs $u(t) = (F(\theta(t), G_i))_{i=1}^N$ follow the dynamics
\begin{equation}
  \frac{du}{dt} = -H(t) (u(t) - \yb), \text{ where } H(t)_{ij} = \Big\langle \frac{\partial F(\theta(t),G_i)}{\partial \theta}, \frac{\partial F(\theta(t),G_j)}{\partial \theta}  \Big\rangle.
\end{equation}
Here, $\yb = (y_i)_{i=1}^N$.
If $\theta$ is sufficiently large (i.e., the network sufficiently wide), then it was shown that the matrix $H(t) \in \R^{N \times N}$ remains approximately constant as a function of $t$. In this case, the neural network becomes approximately a kernel regression \cite{schoelkopf01}. If the parameters $\theta(0)$ are initialized as i.i.d.\ Gaussian, then the matrix $H(0)$ converges to a deterministic kernel matrix $\widetilde{H}$, the \emph{Neural Tangent Kernel}, with closed form regression solution $F_{\widetilde{H}}(G)$.
Given this approximation, one may analyze generalization via kernel learning theory.
\begin{theorem}[\cite{bartlett02rademacher}]
  Given $N$ i.i.d.\ training data points and any loss function $\ell: \R \times \R \to [0,1]$ that is 1-Lipschitz in the first argument with $\ell(y,y)=0$, with probablity $1-\delta$ the population risk of the Graph Neural Tangent predictor is bounded as
  \begin{equation*}
    \risk(F_{\widetilde{H}}) = O\Big( \tfrac{1}{N}\sqrt{\yb^\top \widetilde{H}^{-1} \yb\cdot \tr(\widetilde{H})} + \sqrt{\tfrac{1}{N}\log(1/\delta)} \Big).
  \end{equation*}
\end{theorem}
In contrast to the results in Section~\ref{sec:gen_comp}, the complexity measure $\yb^\top \widetilde{H}^{-1} \yb$ of the target function is data-dependent.
If the target function to be learned follows a simple GNN structure with a polynomial, then this bound can be polynomial:
\begin{theorem}[\cite{du19gntk}]\label{thm:gntk-poly}
  Let $\bar{h}_v = c_v \sum_{u \in \Ns(v) \cup \{v\}}h_u$. If the labels $y_i$, $1 \leq i \leq N$, satisfy
  \begin{equation*}
    y_i = \alpha_1 \sum_{v \in V(G_i)}\beta_1^\top\bar{h}_v + \sum_{l=1}^\infty \alpha_{2l} \sum_{v \in V} \big( \beta_{2l}^\top\bar{h}_v\big)^{2l}
  \end{equation*}
  for $\alpha_k \in \R, \beta_k \in \R^d$, then $\yb^\top \widetilde{H}^{-1} \yb \leq 2 |\alpha_1| \cdot \|\beta_1\|_2 + \sum_{l=1}^\infty \sqrt{2\pi}(2l - 1)|\alpha_{2l}|\cdot \|\beta_{2l}\|_2^{2l}$. With $n = \max_i V(G_i)$, we have $\tr(\widetilde{H}) = O(n^2N)$.  
\end{theorem}

\subsection{Generalization via Algorithmic Alignment}\label{sec:gen_align}
The Graph NTK analysis shows a polynomial sample complexity if the function to be learned is close to the computational structure of the GNN, in a simple way. While this applies to mainly simpler learning tasks, the idea of an ``alignment'' of computational structure carries further.
Recently, there has been growing interest in learning scientific tasks, e.g., given a set of particles or planets along with their location, mass and velocity, predict the next state of the system \cite{battaglia2016interaction,santoro17simple,santoro2018}, and in ``algorithmic reasoning'', e.g., learning to solve combinatorial optimization problems in particular over graphs \cite{cappart2021survey}. In such cases, the target function corresponds to an algorithm, e.g., a dynamic program.

While many neural network architectures have the power to represent such tasks, empirically, they do not learn them equally well from data. In particular, GNNs perform well here, i.e., their architecture encodes suitable \emph{inductive biases} \cite{battaglia18relational,xulz20}. As a concrete example, consider the Shortest Path problem. The computational structure of MPNNs matches that of the Bellman-Ford (BF) algorithm \cite{bellman58sp} very well: both ``algorithms'' iterate, and in each iteration $t$, update the state as a function of the neighboring nodes and edge weights $w(u,v)$:
\begin{equation}\label{eq:shortest}
  \text{BF: }\; d[t][v] = \min_{u \in \mathcal{N}(v)} \; d[t-1][u] + w(u,v) \quad   \text{GNN: }\; h^{(t)}_v = \sum_{u \in \mathcal{N}(v)} \mathrm{MLP}( h^{(t-1)}_u, h^{(t-1)}_v, w(u,v)).
\end{equation}
Hence, the GNN can simulate the BF algorithm if it uses sufficiently many iterations, and if the aggregation function approximates the BF state update (relaxation step). Intuitively, this update is a much simpler function to learn than the full algorithm as a black box, i.e., the GNN encodes much of the algorithmic structure, sparsity and invariances in the architecture. More generally, MPNNs match the structure of many dynamic programs in an analogous way \cite{xulz20}, as long as the updates are permutation invariant or sufficient node identification is provided as input, in light of the results in Section~\ref{sec:representation}. \citet{dudzik22} refine and generalize the relations between GNNs and dynamic programming by using category theory.

The NTK results formalize simplicity by a small function norm in the RKHS associated with the Graph NTK; this can become complicated with more complex tasks and multiple layers. To quantify \emph{structural} match, Xu et al.~\cite{xulz20} define \emph{algorithmic alignment} by viewing a neural network as a structured arrangement of learnable modules -- in a GNN, the (MLPs in the) aggregation functions -- and define complexity via sample complexity of those modules in a PAC-learning framework. Sample complexity in PAC learning is defined as follows: We are given a data sample $\{(x_i,y_i)\}_{i=1}^N$ drawn i.i.d.\ from a distribution $\Ps$ that satisfies $y_i = g(x_i)$ for an underlying target function $g$. Let $f = \mathcal{A}(\{x_i,y_i\}_{i=1}^N)$ be the function output by a learning algorithm $\mathcal{A}$. For a fixed error $\epsilon$ and failure probability $1-\delta$, the function $g$ is $(N, \epsilon, \delta)$-\emph{PAC learnable} with $\mathcal{A}$ if
\begin{equation}
  \mathbb{P}_{x \sim \Ps} \big[ |f(x) - g(x)| < \epsilon\,\big] \geq 1-\delta.
\end{equation}
The \emph{sample complexity} $\Cs_{\As}(g, \epsilon, \delta)$ is the smallest $N$ so that $g$ is $(N,\epsilon,\delta)$-learnable with $\As$.

\begin{definition}[Algorithmic Alignment]
  Let $g$ be a target function and $\mathcal{N}$ a neural network with $M$ modules $\mathcal{N}_i$. The module functions $f_1, ..., f_M$ generate $g$ for $\mathcal{N}$ if, by replacing $\mathcal{N}_i$ with $f_i$, the network $\mathcal{N}$ simulates $g$. Then $\mathcal{N}$  $(N, \epsilon, \delta)$-\emph{algorithmically aligns} with $g$ if (1) $f_1, ..., f_M$ generate $g$ and (2) there are learning algorithms $\mathcal{A}_i$ for learning $f_i$ with $\mathcal{N}_i$, with sample complexity $M \cdot \max_i C_{\mathcal{A}_i}(f_i, \epsilon, \delta) \leq N$. 
\end{definition}
Algorithmic alignment resembles Kolmogorov complexity~\cite{kolmogorov1998tables}. Thus, it can be hard to obtain the optimal alignment  between a neural network and an algorithm. But, \emph{any} algorithmic alignment yields a bound, and any with acceptable sample complexity may suffice.
The complexity of the MLP modules in GNNs may be measured with a variety of techniques. One option is the NTK framework. The module-based bounds then resemble the polynomial bound in Theorem~\ref{thm:gntk-poly}, since both are extensions of \cite{arora19finegrained}. However, here, the bounds are applied at a module level, and not for the entire GNN as a unit. Theorem~\ref{thm:align-generalization} translates these bounds, in a simplified setting, into sample complexity bounds for the full network.
\begin{theorem}[\cite{xulz20}] \label{thm:align-generalization}
Fix $\epsilon$ and $\delta$. Suppose $\left\lbrace G_i, y_i \right\rbrace_{i=1}^N \sim \mathcal{P}$, where $\lvert V(G_i) \rvert \leq n$, and $y_i = g(G_i)$ for some $g$. Suppose $\mathcal{N}_1, ..., \mathcal{N}_M$ are  network $ \mathcal{N}$'s MLP modules  in sequential order of processing. Suppose $\mathcal{N}$ and $g$  $(N, \epsilon, \delta)$-algorithmically align via functions $f_1, ..., f_M$ for a constant $M$. Under the following assumptions, $g$ is  $(N, O( \epsilon), O( \delta))$-learnable by $\mathcal{N}$. 
\vspace{0.05in} \\
  \textbf{a) Sequential learning.} We train $\mathcal{N}_i$'s sequentially: $\mathcal{N}_1$ has input samples $\{\hat{x}_i^{(1)},f_1 (\hat{x}_i^{(1)})\}_{i=1}^N$, with $\hat{x}_i^{(1)}$ obtained from $G_i$. 
For $j > 1$, the input $\hat{x}_i^{(j)}$ for $\mathcal{N}_j$ are the outputs of the previous modules, but labels are generated by the correct functions $f_{j-1}, ..., f_1$ on  $\hat{x}_i^{(1)} $.  \\
 \textbf{b) Algorithm stability.} Let $\mathcal{A}$ be the learning algorithm for the $\mathcal{N}_i$'s. Suppose $f = \mathcal{A}(\left\lbrace x_i, y_i \right\rbrace_{i=1}^N)$, and $\hat{f} = \mathcal{A} ( \left\lbrace \hat{x}_i, y_i \right\rbrace_{i=1}^N) $. For any $x$, $\| f(x) - \hat{f}(x) \| \leq L_0 \cdot \max_i \| x_i - \hat{x}_i \|$, for some $L_0 < \infty$.  \\
 \textbf{c) Lipschitzness.}  The learned functions $\hat{f}_j$ satisfy $ \| \hat{f}_j (x) - \hat{f}_j (\hat{x})  \| \leq L_1 \| x - \hat{x} \|$, for some $L_1 < \infty$. 
\end{theorem}
The big $O$ notation here hides factors including the Lipschitz constants, number of modules and graph size. When measuring module complexity via the NTK, Theorem~\ref{thm:align-generalization} indeed yields a gap in upper bounds between fully connected networks and GNNs in simple cases \cite{xulz20}, supporting empirical results.
While some works use sequential training in experiments \cite{Velic2020Neural}, empirically, better alignment improves learning and generalization even with more common ``end-to-end'' training, i.e., optimizing all parameters simultaneously \cite{xulz20,battaglia18relational}. 

At a general level, these alignment results indicate 
how incorporating expert knowledge, e.g. in terms of algorithmic techniques or physics, into the design of the learning method can improve sample efficiency.

%% file: extrapolation1.tex
\section{Extrapolation}\label{sec:extrapolation}

Section~\ref{sec:generalization} summarizes results for in-distribution generalization, i.e., how well a learned model performs on data from the same distribution $\Ps$ as the training data. Yet, in many practical scenarios, a model is applied to data from a different distribution. A strong case of such a distribution shift is \emph{extrapolation}. It considers the expected loss $\mathbb{E}_{G \sim \Qs}[ \ell(G,g(G), F(G))]$ under a distribution $\Qs$ with  different support, e.g., $\supp(\Qs) \supset \supp(\Ps)$.
For graphs, $\Qs$ may entail graphs of different sizes, different degrees, or with node attributes in different ranges from the training graphs. As no data has been observed in $\supp(\Qs) \setminus \supp(\Ps)$, extrapolation can be ill-defined without stronger assumptions on the task and model class. At least two types of such assumptions have been made. Theoretical results on extrapolation either assume that the graphs have sufficient structural similarity or that the model class is sufficiently restricted to extrapolate accurately. Empirically, while extrapolation has been difficult, several works achieve GNN extrapolation in tasks like predicting the time evolution of physical systems~\cite{battaglia2016interaction}, learning graph algorithms~\cite{Velic2020Neural}, and solving equations~\cite{Lample2020Deep}. 

\textbf{Structural similarity of graphs.} One possibility to guarantee successful extrapolation to larger graphs is to assume sufficient structural similarity between the graphs in $\Ps$ and $\Qs$, in particular, structural properties that matter for the GNN family under consideration.
For spectral GNNs, this assumption has been formalized as the graphs arising from the same underlying topological space, manifold or graphon. Under such conditions, spectral GNNs -- with conditions on the employed filters -- can generalize to larger graphs \cite{ruiz20graphon,ruiz21,levie21transfer}. The underlying structure also ensures similar local structure of the graphs.

For spatial message passing GNNs, whose representations rely on computation trees as local structures (Section~\ref{sec:mpgnn_iso}), an agreement in the distributions of the computation trees in the graphs sampled from $\Ps$ and $\Qs$ is necessary \cite{yehudai21}. This is violated, for instance, if the degree distribution is a function of the graph size, as is the case for random graphs under the Erd\H{o}s-R\'enyi or Preferential Attachment models. The computation tree of depth $t$ rooted at a node $v$ corresponds to the color $c^{(t)}(v)$ assigned by the 1-WL algorithm.
\begin{theorem}[\cite{yehudai21}]\label{thm:extra-overlap}
  Let $\Ps$ and $\Qs$ be finitely supported distributions of graphs. Let $\Ps^t$ be the distribution of colors $c^{(t)}(v)$ over $\Ps$ and similarly $\Qs^{t}$ for $\Qs$. Assume that any graph in $\Qs$ contains a node with a color in $\Qs^t\setminus \Ps^t$. Then, for any graph regression task solvable by a GNN with depth $t$ there exists a GNN with depth at most $t + 3$ that perfectly solves the task on $\Ps$ and predicts an answer with arbitrarily large error on all graphs from $\Qs$.
\end{theorem}
The proof exploits the fact that GNN predictions on nodes only depend on the associated computation tree and that a sufficiently flexible GNN can assign arbitrary target labels to any computation tree \cite{yehudai21,morris19}. I.e., the available information allows for multiple local minima of the empirical risk. A similar result can be shown for node prediction tasks. (A ``sufficiently large'' GNN here means depth at least $t+2$ layers and width $\max\{(\max\deg(G)+1)^t \cdot |C|, 2 \sqrt{|P|}\}$, where the max degree refers to any graph in the support, $|C|$ is the finite number of possible input node attributes and $P$ the set of colors encountered in graphs in the support.)

\textbf{Conditions on the GNN.} If sufficient structural similarity of the input graphs cannot be guaranteed, then further restrictions on the GNN can enable extrapolation to different graph sizes, structures and ranges of input node attributes. If there are no training observations in a certain range of attributes or set of local structures, then the predictions of the learned model depend on the \emph{inductive biases} induced by the model architecture, loss function and training algorithm. Which prediction function, out of multiple fitting functions, a model will choose, depends on these biases. 

\citet{xu21extra} analyze such biases to obtain conditions on the GNN for extrapolation. 
Taking the perspective of algorithmic alignment (Section~\ref{sec:gen_align}), they first analyze how individual module functions, i.e., the MLPs in the aggregation function of a GNN, extrapolate, and then transfer this to the entire GNN. The aggregation functions enter the extrapolation regime, e.g., if the node attributes, node degrees or computation trees are different under $\Qs$ compared to $\Ps$, as they determine the inputs to the aggregations.
The following theorem states that, sufficiently far away from $\supp(\Ps)$, MLPs implement directionally linear functions.
\begin{theorem}[\cite{xu21extra}] \label{thm:non-linear}
Suppose we train a two-layer MLP $f: \mathbb{R}^d \rightarrow \mathbb{R}$ with ReLU activation functions with squared loss in the Neural Tangent Kernel (NTK) regime. For any direction $v \in \mathbb{R}^d$, let $x_0 = tv$. As $t \rightarrow \infty$,  $f(x_0 + h v) - f(x_0)  \rightarrow \beta_{v} \cdot h $ for any $h > 0$, where $\beta_{v}$ is a constant linear coefficient. Moreover, given $\epsilon > 0$,  for $t = O(\frac{1}{\epsilon})$, we have $\vert \frac{f(x_0 + h v) - f(x_0)}{h} -  \beta_{v} \vert < \epsilon$.
\end{theorem}
The linear function and the constant terms in the convergence rate depend on the training data and the direction $v$. The proof of Theorem~\ref{thm:non-linear} relies on the fact that a neural network in the NTK regime learns a minimum-norm interpolation function \cite{jacot18,arora19finegrained,arora19exact}. Although Theorem~\ref{thm:non-linear} uses a simplified setting of a wide 2-layer network, similar results hold empirically for more general MLPs \cite{xu21extra}.

To appreciate the implications of this result in the context of GNNs, consider the example of Shortest Path in Equation~\eqref{eq:shortest}. For the aggregation function to mimic the Bellman-Ford algorithm, the MLP must approximate a nonlinear function. But, in the extrapolation regime, it implements a linear function and therefore is expected to not approximate Bellman Ford well any more. Indeed, empirical works that successfully extrapolate GNNs for Shortest Path use a different aggregation function of the form \cite{battaglia18relational, Velic2020Neural}
\begin{align}
\label{eq:min-gnn}
h_u^{(t)} = \max_{v \in \mathcal{N}(u)} \text{MLP}^{(t)} \big( h_u^{(t - 1)}, h_v^{(t - 1)}, w_{(v, u)} \big).
\end{align}
Here, the nonlinear parts do not need to be learned, allowing to extrapolate with a linear learned MLP. More generally, the directionally linear extrapolation suggests that (1) the architecture or (2) the input encoding should be set up such that the target function can be approximated when MLPs learn linear functions (\emph{linear algorithmic alignment}). An example for (2) may be found in forecasting physical systems, e.g., predicting the evolution of $n$ objects in a gravitational system, and the node (object) attributes are mass, location and velocity at time $t$. The position of an object at time $t+1$ is a nonlinear function of the attributes of the other objects. When encoding the nonlinear function as transformed edge attributes, the function to be learned becomes linear. Many empirical works that successfully extrapolate implement the idea of linear algorithmic alignment \cite{xu21extra,trask18,johnson17,yi18,mao19,cranmer20}.

Finally, the geometry of the training data also plays an important role. \cite{xu21extra} show empirical results and initial theoretical results for learning max-degree, that, even with linear algorithmic alignment, sufficient diversity in the training data is needed to identify the correct linear functions. These data conditions are weaker than those implied by Theorem~\ref{thm:extra-overlap}, due to the linear algorithmic alignment assumption.

For the case when the target test distribution $\Qs$ is known, \citet{yehudai21} propose approaches for combining elements of $\Ps$ and $\Qs$ to enhance the range of the data seen by the GNN.

%% file: disc.tex
\section{Conclusion}
This survey summarized three main topics in theoretically understanding GNNs: representation and approximation, generalization, and extrapolation. As GNNs are an active research area, many results could not be included. E.g., we focused on MPNNs and main ideas for higher-order GNNs, but neglected spectral GNNs, which closely relate to ideas in graph signal processing. Other emergent topics include adversarial robustness, optimization behavior of the empirical risk and its improvements, and computational scalability and approximations. Overall, GNNs have a rich set of mathematical connections, a selection of which was covered here.

Many questions remain. Regarding approximation capabilities, the limitations of MPNNs have motivated powerful higher-order GNNs. However, these are still computationally expensive. What efficiency is theoretically possible? Moreover, most applications may not require full graph isomorphism power, or $k$-WL power for large $k$. What other measures of representational power make sense? Do they allow better and sharper complexity results? Initial works consider subgraph counting as a benchmark task \cite{chen20count,tahmasebi21}.

The generalization results so far need to use simplifications in the analysis, similar to most theoretical analyses of deep learning. To what extent can they be relaxed? Do more specific tasks or graph classes allow sharper results? Which modifications of GNNs would allow them to generalize better, and how do higher-order GNNs generalize?
Similar questions pertain to extrapolation and reliability under distribution shifts, a topic that has been studied even less than GNN generalization.

In general, revealing further mathematical connections may enable the design of richer models and enable a more thorough understanding of GNNs' learning abilities and limitations, and eventual improvements.